\documentclass{article} 
\usepackage{nips14submit_e,times}
\usepackage{graphicx}
\usepackage{amsmath,amssymb} 
\usepackage{color}

\usepackage{titlesec}
\titleformat{\subsubsection}[runin]{\normalfont\bfseries}{}{}{}[]

\usepackage{xspace}
\usepackage[colorlinks=true]{hyperref}
\usepackage[all]{hypcap}
\usepackage[stable]{footmisc}

\usepackage{multirow}
\usepackage{floatrow}
\newfloatcommand{capbtabbox}{table}[][\FBwidth]
\usepackage{subcaption}

\usepackage{tikz}
\usetikzlibrary{spy}
\usetikzlibrary{shadows}
\usetikzlibrary{calc}
\usetikzlibrary{fit}
\usetikzlibrary{positioning}
\usetikzlibrary{backgrounds}
\usetikzlibrary{shapes.misc}

\usepackage{pgfplots}
\pgfplotsset{compat=newest} 
\pgfplotsset{plot coordinates/math parser=false}

\tikzset{
  spy using overlaysshadow/.style={
    spy scope={#1,
      every spy on node/.style={
        rectangle,
        fill, fill opacity=0.5, text opacity=1
      },
      every spy in node/.style={
        rectangle, drop shadow={opacity=0.5,fill=black},
        fill=white, draw, ultra thick, cap=round
      }
    }
  },
  blackbox/.style={
    rounded corners,
    minimum size=2cm,
    top color=black!20,
    bottom color=black!50,
  },
  container/.style={
    rounded corners,
    fill=black!10,
    inner sep=0.2cm,
  }
}

\newlength\figureheight
\newlength\figurewidth 

\newcommand{\fig}[1]{Figure~\ref{fig:#1}}
\newcommand{\sect}[1]{Section~\ref{sect:#1}}
\newcommand{\tab}[1]{Table~\ref{tab:#1}}

\newcommand{\NQ}{\ensuremath{N^4}\xspace}
\renewcommand{\P}{\ensuremath{{\cal P}}\xspace}
\newcommand{\A}[1]{\ensuremath{{\mathbf A}(#1)}\xspace}
\newcommand{\B}[1]{\ensuremath{{\mathbf B}(#1)}\xspace}
\newcommand{\Seg}[1]{\ensuremath{{\mathbf S}(#1)}\xspace}
\newcommand{\Full}{\ensuremath{\mathbf F}\xspace}
\newcommand{\CNN}{\ensuremath{\text{CNN}}\xspace}
\newcommand{\NNB}{\ensuremath{\text{NNB}}\xspace}
\newcommand{\PCA}{\ensuremath{\text{PCA}}\xspace}

\title{$N^4$-Fields: Neural Network Nearest Neighbor Fields for Image Transforms} 

\author{
Yaroslav Ganin, Victor Lempitsky \\
Skolkovo Institute of Science and Technology (Skoltech) \\
\texttt{\{ganin, lempitsky\}@skoltech.ru} \\
}

\nipsfinalcopy 

\begin{document}

\maketitle

\begin{abstract}
We propose a new architecture for difficult image processing operations, such as natural edge detection or thin object segmentation. The architecture is based on a simple combination of convolutional neural networks with the nearest neighbor search. 

We focus our attention on the situations when the desired image transformation is too hard for a neural network to learn explicitly. We show that in such situations, the use of the nearest neighbor search on top of the network output allows to improve the results considerably and to account for the underfitting effect during the neural network training. The approach is validated on three challenging benchmarks, where the performance of the proposed architecture matches or exceeds the state-of-the-art.
\end{abstract}

\section{Introduction}

Deep convolutional neural networks (CNNs) \cite{LeCun89} have recently achieved a breakthrough in a variety of computer vision benchmarks and are attracting a very strong interest within the computer vision community. The most impressive results have been attained for image \cite{Krizhevsky12} or pixel \cite{Ciresan12} classification results. The key to these results was the sheer size of the trained CNNs and the power of modern GPU used to train those architectures.

Here, we demonstrate that convolutional neural networks can achieve state-of-the-art results for sophisticated image processing tasks. The complexity of these tasks defies the straightforward application of CNNs, which perform reasonably well, but clearly below state-of-the-art.

Here, we demonstrate that by pairing convolutional networks with a simple non-parametric transform based on nearest-neighbor search state-of-the-art performance is achievable. This is demonstrated on three challenging and competitive benchmarks (edge detection on Berkeley Segmentation dataset \cite{Arbelaez11}, edge detection on the NYU RGBD dataset\cite{Silberman11}, retina vessel segmentation on the DRIVE dataset \cite{Staal04}). All the results are obtained with the same meta-parameters, such as the configuration of a CNN, thus demonstrating the universality of the proposed approach.

The two approaches, namely convolutional {\bf N}eural {\bf N}etworks and {\bf N}earest {\bf N}eighbor search are applied sequentially and in a patch-by-patch manner, hence we call the architecture \emph{\NQ-fields}. At test time, an \NQ-field first passes each patch through a CNN. For a given patch, the output of the first stage is a low-dimensional vector corresponding to the activations of the top layer in the CNN. At the second stage we use the nearest neighbor search within the CNN activations corresponding to patches sampled from the training data. Thus, we retrieve a patch with a known pixel-level annotation that has a similar CNN activation, and transfer its annotation to the output. By averaging the outputs of the overlapping patches, the transformation of the input image is obtained. 

Below, we first review the related works (\sect{related}), describe the proposed architecture and the associated training procedures in detail (\sect{main}), and discuss the results of applying it on sample problems (\sect{results}). We conclude with a short discussion of the merits and the potential of the proposed approach (\sect{discussion}).

\begin{figure}[t]
\centering
\setlength\figureheight{3.7cm}
\begin{tabular}{cccc}
\includegraphics[height=\figureheight]{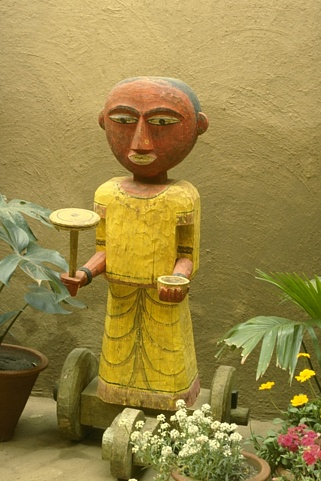}&
\includegraphics[height=\figureheight]{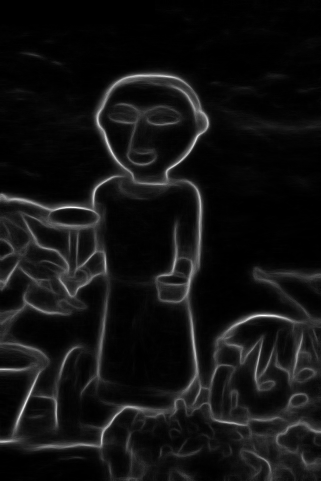}&
\includegraphics[height=\figureheight]{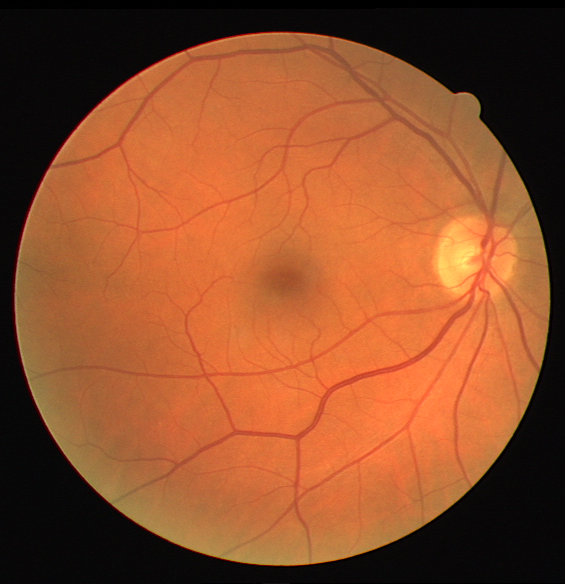}&
\includegraphics[height=\figureheight]{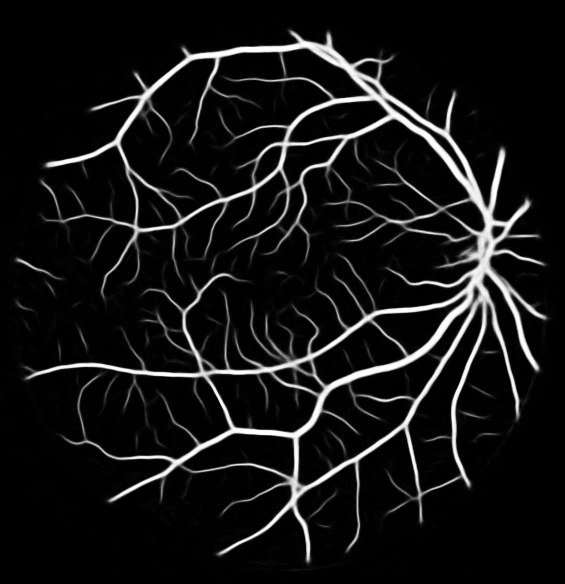}\vspace{-2mm}\\
\end{tabular}
\caption{\NQ-Fields can be applied to a range of complex image processing tasks, such as natural edge detection (left) or vessel segmentation (right). The proposed architecture combines the convolutional neural networks with the nearest neighbor search and is generic.  E.g.\ it achieves state-of-the-art performance on standard benchmarks for these two rather different applications with very little customization or parameter tuning.\vspace{-2mm}}
\label{fig:teaser}
\end{figure}

\section{Related work}
\label{sect:related}

There is a very large body of related approaches, as both neural networks and nearest neighbor methods have been used heavily as components within image processing systems. Here, we only review several works that are arguably most related to ours.

\subsubsection*{Neural networks for image processing.}
The use of neural networks for image processing goes back for decades \cite{Egmont02}. Several recent works have investigated large-scale training of deep architectures for complex edge detection and segmentation tasks. Thus, Mnih and Hinton~\cite{Mnih10} have used a cascade of two deep networks to segment roads in aerial images, while Shulz et al.~\cite{Schulz12} use CNNs to perform semantic segmentation on standard datasets. Kivinen et al.~\cite{Kivinen14} proposed using unsupervised features extraction via deep belief net extension of mcRBM~\cite{Ranzato10} followed by supervised NN training for boundary prediction in natural images. State-of-the-art results on several semantic segmentation datasets were obtained by Farabet et al.~\cite{Farabet13} by using a combination of a CNN classifier and superpixelization-based smoothing. Finally, a large body of work, e.g.~\cite{Jain07,Ciresan12} simply frame the segmentation problem as patch classification, making generic CNN-based classification easily applicable and successful. Below, we compare \NQ-fields against such baseline and find them to achieve better results for our applications.

Another series of works \cite{Jain08,Burger12} investigate the use of convolutional neural networks for image denoising. In this specific application, CNNs wildly benefit from virtually unlimited training data that can be synthesized and have a high realism.

Finally, neural networks have been applied for descriptor learning, which resembles the way they are used within \NQ-fields. Thus, Chopra~et~al.~\cite{Chopra05} introduced a general scheme for learning CNNs that map input images to multi-dimensional descriptors, suitable among other things for nearest neighbor retrieval or similarity verification. The learning in that case is performed on a large set of pairs of matching images. \NQ-fields is thus different from this group of the approaches in terms of their purpose (image processing) and the type of the training data (annotated images).

\subsubsection*{Non-parametric approaches to image processing.}
Nearest neighbor methods have been applied to image processing with a considerable success. Most methods use nearest neighbor relations within the same image, e.g.\ Dabov et al.~\cite{Dabov08} for denoising or Criminisi et al.~\cite{Criminisi04} for inpainting. More related to our work, Freeman et al.~\cite{Freeman00} match patches in a given image to a large dataset of patches from different images, to infer the missing high-frequencies and to achieve super-resolution. All these works use the patches themselves or their band-passed versions to perform the matching. 

Another popular non-parametric framework to perform operations with patches are random forests. Our work was in many ways inspired by the recent impressive results in Doll\'{a}r et al.~\cite{Dollar13}, where random forests are trained on patches with structured annotations. Their emphasis is on natural edge detection, and their system represent the state-of-the-art for this task. \NQ-fields match the accuracy of \cite{Dollar13} for natural edge detection, and perform considerably better for the task of vessel segmentation in micrographs, thus demonstrating the ability to adapt to new domains.

\section{\NQ-Fields}
\label{sect:main}

\subsection{Architecture}
\begin{figure}[t]
\centering
\scalebox{0.75}{\begin{tikzpicture}[remember picture]
  \def\input_width{481}
  \def\desired_width{6cm}
  \def\spy_size{68 * (\desired_width / \input_width)}
  \def\output_patch_size{16 * (\desired_width / \input_width)}
  
  \begin{scope}[spy using overlaysshadow={green,magnification=2,
                size=\spy_size, connect spies}]
    \node [anchor=south west,inner sep=0] (input) at (0,0)
      {\pgfimage[width=\desired_width]{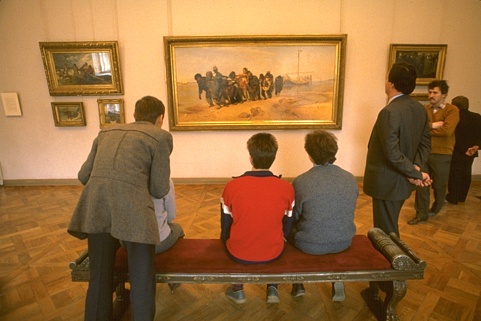}};
    \begin{scope}[x={(input.south east)}, y={(input.north west)}]
      \coordinate (patch_center) at (0.525,0.5);
      \coordinate (spy_center) at (0.675,0.65);
      \spy on (patch_center) in node (spy_node) at (spy_center);
    \end{scope}
  \end{scope}
  
  \begin{scope}[anchor=north west,shift={($(input.north east) + (0.2cm,0)$)},
                local bounding box=nn_and_features]
    \node (neural_net) [blackbox] {ConvNet};
    \draw (spy_node.east) edge [-latex,shorten >=1pt,ultra thick,green,bend left=10] 
      (neural_net.west);
    
    \node (features) [right=0.2cm of input.south east,anchor=south west,container] {
      \begin {tikzpicture} [inner sep=0pt]
        \pgfmathsetseed{42}
        \node (components) at (0,0) {
          \begin{tikzpicture}
            \foreach \bin in {0,1,...,7} {
              \filldraw [sharp corners,draw=blue,fill=blue!50] 
                ($ (\bin*0.2, 0) $) rectangle +($ (0.2, rand*0.2+0.2) $);
            }
          \end{tikzpicture}
         };
         \node [below=2pt of components] {\footnotesize Features};
      \end{tikzpicture}
    };
    \draw (neural_net.south) edge [-latex, shorten >=1pt, ultra thick, green]
      (features.north);
  \end{scope}
  
  \node (dict) [right=0.8cm of nn_and_features.east,container] {
    \begin {tikzpicture} [inner sep=0pt]
      \pgfmathsetseed{42}
      \node (dict_internal) at (0,0) {
        \begin{tikzpicture}[inner sep=0pt]
          \foreach \y in {1,...,4} {
            \node (patch_\y) [anchor=south] at (0, \y) 
              {\pgfimage{./Images/pipeline_patch_\y.png}};
            \node (feats_\y)[right=0.1cm of patch_\y] {
              \begin{tikzpicture}
                \foreach \bin in {0,1,...,7} {
                  \filldraw [sharp corners,draw=blue,fill=blue!50] 
                    ($ (\bin*0.2, 0) $) rectangle +($ (0.2, rand*0.2+0.2+0.05) $);
                }
              \end{tikzpicture}
            };
          }
        \end{tikzpicture}
      };
      \node [below=5pt of dict_internal] {\footnotesize Dictionary};
    \end{tikzpicture}
  };

  \node (nearest) [fit=(patch_1) (feats_1),draw,rectangle,ultra thick,green] {};
  \draw (features.east) edge [-latex,shorten >=1pt,ultra thick,green,bend right=10] 
      (nearest.west);

  \begin{scope}[anchor=west,shift={($(dict.east) + (0.2cm,0)$)}]
    \node [anchor=west,inner sep=0] (output) at (0,0)
      {\pgfimage[width=\desired_width]{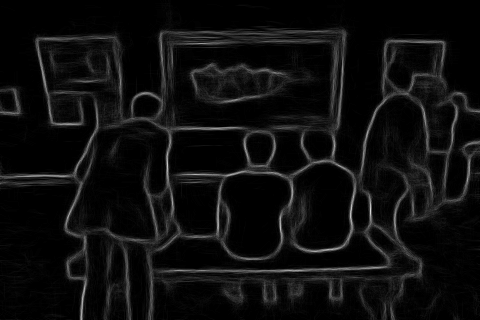}};
    \begin{scope}[shift={(output.south west)},
                  x={(output.south east)},y={(output.north west)}]
      \node (output_patch) 
        [draw,green,ultra thick,
         minimum size=\output_patch_size,anchor=center] at (0.525,0.5) {};
    \end{scope}
  \end{scope}

  \draw (nearest.east) edge [-latex,shorten >=1pt,ultra thick,green,bend right=10] 
      (output_patch.west);
\end{tikzpicture}\vspace{-4mm}}
\caption{The \NQ architecture for natural edge detection. The input image is processed patch-by-patch. An input patch is first passed through a pretrained convolutional neural network (CNN). Then, the output of the CNN is matched against the dictionary of sample CNN outputs that correspond to training patches with known annotations. The annotation corresponding to the nearest neighbor is transferred to the output. Overall, the output is obtained by averaging the overlapping transferred annotations. \vspace{-2mm}}
\label{fig:main_arch}
\end{figure}
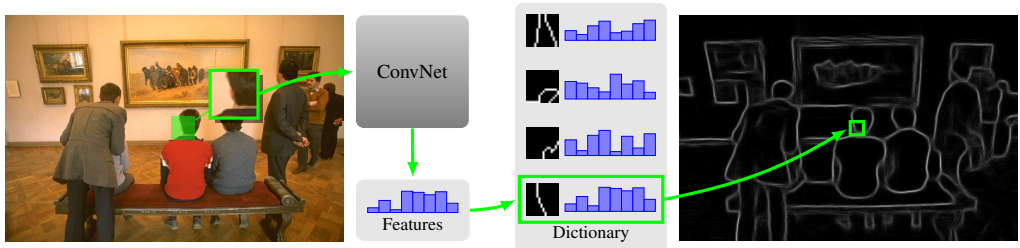

We start by introducing the notation, and discussing the way our architecture is applied to images. The \NQ-Fields transform images patch-by-patch. Given an image transform application, we wish to map a single or multi-channel (e.g. RGB) image patch \P of size $M\times M$ to a segmentation, an edge map, or some other semantically-meaningful annotation \A{\P}, which in itself is a single or multi-channel image patch of size $N\times N$. We take $N$ to be smaller than $M$, so that \A{\P} represents a desired annotation for the central part of \P. 

Given the annotated data, we learn a mapping $\Full$ that maps patches to the desired annotations. At test time, the mapping is applied to all image patches and their outputs are combined by averaging, thus resulting in an output image. The output of processing for a pixel $p=(x,y)$ is the average of the outputs of $N^2$ patches that contain this pixel. More formally, the output of the mapping on the input image $I$ is defined as:
\begin{equation}
\Full(I)[x,y] = \frac{1}{N^2} \sum_{i,j: |i-x| \le N/2 \atop |j-y| \le N/2} \Full\left( I(i,j|M) \right)[x-i,y-i]\,, \label{eq:transform}
\end{equation}
where $\Full(I)[x,y]$ denotes the value of image transform at pixel $(x,y)$, $I(i,j|M)$ denotes the image patch of size $M\times M$ centered at $(i,j)$, and $\Full\left( I(i,j|M) \right)[x-i,y-i]$ is a pixel in the output patch at the position $(x-i,y-j)$ assuming the origin in the center of the patch. 

Obviously, the accuracy of the transform depends on the way the transform $\Full$ is defined and learned. Convolutional neural networks (CNNs) provide a generic architecture for learning functions of the multi-channel images and patches exploiting the translational invariance properties of natural images. The direct approach is then to learn a mapping $\P \to \A{\P}$ in the form of a CNN.
In practice, we found the flexibility of CNNs to be insufficient to learn the corresponding mapping even when a large number of layers with large number of parameters are considered. For complex transforms, e.g.\ natural edge detection, we observe a strong underfitting during the training, which results in a suboptimal performance at test time. 

Convolutional neural network can be regarded as a parametric model, albeit with a very large number of parameters. A straightforward way to increase the fitting capacity of the mapping is to consider a non-parametric model. We thus combine a simple non-parametric mapping (nearest neighbor) and a complex parametric mapping (convolutional neural network). The input patch \P is first mapped to an intermediate representation $\CNN(\P;\Theta)$, where $\Theta$ denotes the parameters of the CNN. The output $\CNN(P;\Theta)$ of the CNN mapping is then compared to a dictionary dataset of CNN outputs, computed for $T$ patches $\P_1,\P_2,\dots,\P_T$ taken from the training images, and thus having known annotations $\A{\P_1},\A{\P_2},\dots,\A{\P_T}$. The input patch is then assigned the annotation from the dictionary patch with the closest CNN output, i.e. $\A{P_k}$, where $k = \arg\min_{i=1}^T ||\CNN(P_i)-\CNN(P)||$ (\fig{main_arch}). If we denote such nearest neighbor mapping as $\NNB$, then the full two-stage mapping is defined as:
\begin{equation}
\Full(\P) = \NNB\left( \CNN(\P;\,\Theta)\, |\, \{(\CNN(\P_i;\Theta);\A{\P_i})|i=1..T\} \right)\,,
\end{equation}
where $\NNB(x\,|\,\{(a_i|b_i\})$ denotes the nearest-neighbor transform that maps $x$ to $b_i$ corresponding to $a_i$ that is closest to $x$.
In our experiments, the dimensionality of the intermediate representation (i.e. the space of CNN outputs) is rather low (16 dimensions), which makes nearest neighbor search reasonably easy.

In the experiments, we observe that such a two-stage architecture can successfully rectify the underfitting effect of the CNN and result in better generalization and overall transform quality compared to single stage architectures that include either CNN alone or nearest neighbor search on hand-crafted features alone.

\subsection{Training}
\label{sect:main_training}

The training procedure for an \NQ-field requires learning the parameters $\Theta$ of the convolutional neural network. Note, that the second stage (nearest neighbor mapping) does not require any training apart from sampling $T$ patches from the training images.

The CNN training is performed in a standard supervised way on the patches drawn from the training images $I_1,I_2,\dots I_R$. For that, we define the surrogate target output for each input patch. Since for each training patch \P, the desired annotation \A{\P} is known, it is natural to take this annotation itself as such a target (although other other variants are possible as described in \sect{main_impl_training_cnn}), i.e.\ to train the network on the input-output pairs of the form $(\P,\A{\P})$. However, such output can be rather high-dimensional (when the output patch size is large) and can vary discontinuously even when the input patch is disturbed or jittered a little, in particular when our model applications of edge detection or thin object segmentations are considered. To address both problems, we perform dimensionality reduction of the output annotations using PCA. Experimentally, we found that the target dimensionality can be taken rather small, e.g.\ 16 dimensions for $16\times16$ patches.

Thus, the overall training process includes the following steps:
\begin{enumerate}
\item Learn the PCA projection on a subset of $N \times N$ patches extracted from the training image annotations.
\item Train the convolutional neural network on the input-output pairs \\$\{(\P,\PCA(\A{\P})\}$ sampled from the training images.
\item Construct a dictionary $\{(\CNN(\P_i;\Theta);\A{\P_i})|i=1..T\}$ by drawing $T$ random patches from the training images and passing them through the trained network.
\end{enumerate}

After the training, the \NQ-field can be applied to new images as discussed above.

\subsection{Implementation details}

\subsubsection*{Training the CNN.}
\label{sect:main_impl_training_cnn}
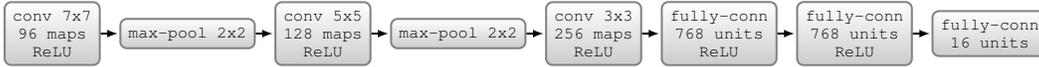
\begin{figure}[t]
\centering
\scalebox{0.62}{\begin{tikzpicture}[node distance=4mm,
terminal/.style={
align=center,
rectangle,minimum size=6mm,rounded corners,
inner sep=5pt,
very thick,draw=black!50,
top color=white,bottom color=black!20,
font=\ttfamily}]
  \node (conv1) [terminal] {conv 7x7\\96 maps\\ReLU};
  \node (pool1) [terminal,right=of conv1] {max-pool 2x2};
  \node (conv2) [terminal,right=of pool1] {conv 5x5\\128 maps\\ReLU};
  \node (pool2) [terminal,right=of conv2] {max-pool 2x2};
  \node (conv3) [terminal,right=of pool2] {conv 3x3\\256 maps\\ReLU};
  \node (fc4) [terminal,right=of conv3] {fully-conn\\768 units\\ReLU};
  \node (fc5) [terminal,right=of fc4] {fully-conn\\768 units\\ReLU};
  \node (fc6) [terminal,right=of fc5] {fully-conn\\16 units};
  
  \path (conv1) edge[-latex,shorten >=1pt,very thick] (pool1);
  \path (pool1) edge[-latex,shorten >=1pt,very thick] (conv2);
  \path (conv2) edge[-latex,shorten >=1pt,very thick] (pool2);
  \path (pool2) edge[-latex,shorten >=1pt,very thick] (conv3);
  \path (conv3) edge[-latex,shorten >=1pt,very thick] (fc4);
  \path (fc4) edge[-latex,shorten >=1pt,very thick] (fc5);
  \path (fc5) edge[-latex,shorten >=1pt,very thick] (fc6);
\end{tikzpicture}\vspace{-4mm}}
\caption{The CNN architecture used in our experiments. See \sect{main_impl_training_cnn} for details.\vspace{-2mm}}
\label{fig:main_cnn_arch}
\end{figure}

We use the heavily modified CNN framework\footnote{\url{https://code.google.com/p/cuda-convnet/}} written by Alex Krizhevsky. It features an efficient GPU implementation of forward and backward propagation and is designed to be easily customizable and extendable thus fits our purposes nicely. The base architecture that was used in our experiments is loosely inspired by \cite{Krizhevsky12}. It is comprised of the layers shown in \fig{main_cnn_arch}. We also tried a dozen of other CNN designs (deeper ones and wider ones) but the performance always stayed roughly the same which suggests that our system is somewhat insensitive to the choice of the architecture given the sufficient number of free parameters. 

The model was trained on $ 34 \times 34 $ patches extracted at randomly sampled locations of the training images. Each patch is preprocessed by subtracting the per-channel mean (across all images). Those patches are packed into mini-batches of size 128 and presented to the network. The initial weights in the CNN are drawn from Gaussian distribution with zero mean and $ \sigma = 10^{-2} $. They are then updated using stochastic gradient descent in conjunction with a momentum term set to 0.9. The starting learning rate $ \eta $ is set to $ 10^{-1} $ (below in \sect{results} we introduce an alternative target function which demands smaller initial $ \eta = 10^{-3} $). As commonly done, we anneal $ \eta $ throughout training when the validation error reaches its plateau.

As the amount of the training data was limited, we observed overfitting (validation error increasing, while training error decreasing) alongside underfitting (training error staying high). To reduce overfitting, we enrich the training set with various artificial transformations of input patches such as random rotations and horizontal reflections. Those transformations are computed on-the-fly during the training procedure. Although batch generation is not free in terms of execution time, it is run in the background hence it does not stall the training pipeline.

Along with data augmentation we apply two regularization techniques which have become quite common for CNNs, namely dropout \cite{Hinton12} (we randomly discard half of activations in the first two fully-connected layers) and $ \ell_2 $-norm restriction of the filters in the first layer \cite{Hinton12,Zeiler12}.

\subsubsection*{Testing procedure.}
At test time we want to calculate activations for patches centered at all possible locations within input images. A naive approach would be to apply a CNN in the sliding window fashion (separate invocation for each location). However this solution may be computationally expensive especially in case of deep architectures. Luckily it is rather easy to avoid redundant calculations and to make dense applications efficient by feeding the network with a sequence of shifted test images \cite{Sermanet14}.    

After neural codes for all patches are computed, nearest-neighbors search is done by means of $ k $-d trees provided as a part of VLFeat package \cite{Vedaldi08vlfeat}. We use default settings except for maximum number of comparisons which we set to 30.

Our proof-of-concept implementation runs reasonably fast taking about 6 seconds to process an image of size $ 480 \times 320 $, although we were not focusing on speed. Computational performance may be brought closer to the real-time by, for example, applying the system in a strided fashion \cite{Dollar13} and finding a simpler design for the CNN.

\subsubsection*{Multi-scale operation.}
Following the works \cite{Dollar13,Ren12} we apply our scheme at different scales. For each input image we combine detections produced for original, half and double resolutions to get the final output. While various blending strategies may be employed, in our case even simple averaging gave remarkably good results.

\subsubsection*{Committee of \NQ-fields.} 
CNNs are shown \cite{Krizhevsky12,Sermanet14,Ciresan12} to perform better if outputs of multiple models are averaged. We found that this technique works quite well for our system too. One rationale would be that different instances of the neural network produce slightly different neural codes hence nearest-neighbor search may return different annotation patches for the same input patch. In practice we observe that averaging amplifies relevant edges and smooths the noisy regions. The latter is especially important for the natural edge detection benchmarks, as the output of \NQ-fields is passed through the non-maximum suppression.

\section {Experiments}
\label{sect:results}

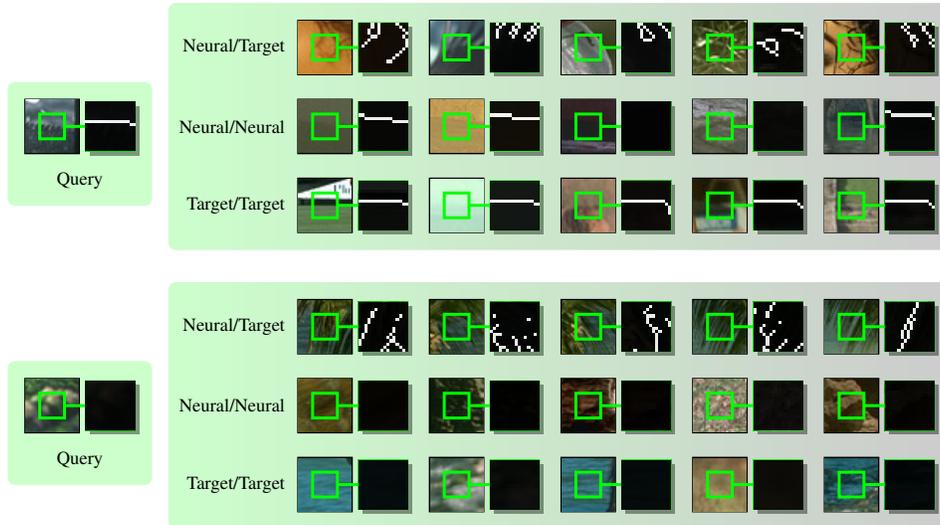
\begin{figure}[t]
  \centering
  \scalebox{0.7}{\tikzset{
  fitting node/.style={
    inner sep=0pt,
    fill=none,
    draw=none,
    reset transform,
    fit={(\pgf@pathminx,\pgf@pathminy) (\pgf@pathmaxx,\pgf@pathmaxy)}
  },
  reset transform/.code={\pgftransformreset}
}
\makeatother

\tikzset{
  every fit/.append style=text badly centered,
  spy using overlaysshadow/.style={
    spy scope={#1,
      every spy on node/.style={
        rectangle,
        draw,
        ultra thick
      },
      every spy in node/.style={
        rectangle, drop shadow={opacity=0.4,fill=black},
        fill=white, draw, cap=round, inner sep=0pt
      },
      spy connection path={\draw[ultra thick] (tikzspyonnode) -- (tikzspyinnode);}
    }
  },
  blackbox/.style={
    rounded corners,
    minimum size=2cm,
    top color=black!20,
    bottom color=black!50,
  },
  container/.style={
    rounded corners,
    fill=black!10,
    inner sep=0.2cm,
  }
}

\newcommand{\neighbourpic}[2]{%
\begin{tikzpicture}[anchor=center]
  \def\inputwidth{34}
  \def\desiredwidth{1cm}
  \dimendef\spysize=0
  \pgfmathsetlength{\spysize}{32 * (\desiredwidth / \inputwidth)}
  
  \begin{scope}[spy using overlaysshadow={green,magnification=2,size=\spysize, connect spies}]
      \node [anchor=south west,inner sep=0,draw=black,ultra thick] (input) at (0,0)
        {\pgfimage[width=\desiredwidth]{#1}};
    \begin{scope}[x={(input.south east)}, y={(input.north west)}]
      \coordinate (patch_center) at (0.5,0.5);
      \coordinate (spy_center) at (1.1,0.5);
      \spy on (patch_center) in node [anchor=west] (spy_node) at (spy_center);
    \end{scope}
  \end{scope}

  \dimendef\annsize=1
  \pgfmathsetlength{\annsize}{\spysize - \pgflinewidth}
  \node [anchor=center,inner sep=0,opacity=0.9] (annotation) at (spy_node.center)
    {\pgfimage[width=\annsize]{#2}};
\end{tikzpicture}%
}%

\pgfdeclarelayer{background}
\pgfdeclarelayer{foreground}
\pgfsetlayers{background,main,foreground}

\newcommand{\neighbourbar}[1]{%
\begin{tikzpicture}
  \def\labels{{"Target/Target", "Neural/Neural", "Neural/Target"}}

  \begin{scope}[local bounding box=nn_scope]
    \foreach \j [evaluate=\j as \jmo using \j - 1] in {1,...,3} {
      \foreach \i [evaluate=\i as \imo using \i - 1] in {1,...,5} {
        \node (nn_\j_\i) at (\imo * 2.5cm, \jmo * 1.5cm) {
          \neighbourpic
            {./Images/nearest/nn_in_#1_\j_\i.png}
            {./Images/nearest/nn_an_#1_\j_\i.png}
        };
      }
      \node (nn_label_\j) [anchor=east,left=0.1 cm of nn_\j_1,inner sep=0] 
        {\pgfmatharray{\labels}{\jmo}\pgfmathresult};
    }
  \end{scope}

  \begin{pgfonlayer}{background}
    \node (nn_container) 
      [container,left color=green!20,right color=black!20,fit={(nn_scope)}] {};
  \end{pgfonlayer}

  \begin{scope}[local bounding box=query_scope]
    \node [anchor=east,left=0.5cm of nn_container.west] (query) {
      \neighbourpic
        {./Images/nearest/in_#1.png}
        {./Images/nearest/an_#1.png}
    };
    \node [below=0.1cm of query] {Query};
  \end{scope}

  \begin{pgfonlayer}{background}
    \node (query_container) [container,fill=green!20,fit={(query_scope)}] {};
  \end{pgfonlayer}
\end{tikzpicture}%
}

\renewcommand{\arraystretch}{1.0}
\begin{tabular}{c}
\neighbourbar{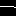}\\[0.5cm]
\neighbourbar{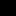}
\end{tabular}
\renewcommand{\arraystretch}{1.0}\vspace{-2mm}}
  \caption{Examples of nearest neighbors. On the left an image patch and its ground truth annotation are shown. Right panels contain the results of the nearest neighbor searches for different combinations of query/dictionary encoding. {\it "Neural"} encoding corresponds to the top-layer activations $\CNN(\P)$ of the CNN trained to produce {\it "Target"} codes $\PCA(\A{\P})$. It can be seen that matches obtained using neural codes on both sides are more adequate than the ones retrieved in the Neural/Target setting, which provide a poor fit in terms of retrieving patches with similar annotations. Thus, there is a gap between the learned neural codes and the target PCA codes (underfitting during CNN training). However, the use of nearest neighbor search allows to overcome this gap and to match patches with appropriate annotations (e.g.\ derived from the 1st nearest neighbor as in all our experiments). See \sect{main_training} and \sect{results_bsds500} for more details. \vspace{-2mm}}
  \label{fig:nearest}
\end{figure}

We evaluate our approach on three datasets. Within two of them (BSDS500 and NYU RGBD), the processing task is to detect natural edges, and in the remaining case (DRIVE) the task is to segment thin vessels in retinal micrographs. Across the datasets, we provide comparison with baseline methods, with the state-of-the-art on those datasets, illustrate the operation of the method, and demonstrate characteristic results.

\subsubsection* {CNN baselines.} 
All three tasks correspond to binary labeling of pixels in the input photographs (boundary/not boundary, vessel/no vessel). It is therefore natural to compare our approach to CNNs that directly predict pixels' labels. Given the input patch a CNN can produce a decision either for the single central pixel or for multiple pixels (e.g. central patch of size $ 16 \times 16 $) hence we have two {\em CNN baselines}. We call them {\em CNN, central} and {\em CNN, patch} respectively. Each of the CNNs has the same architecture as the CNN we use within \NQ-fields, except that the size of the last layer is no longer 16 but equals the number of pixels we wish to produce predictions for (i.e.\ 1 for CNN central and 256 for CNN patch). At test time, we run the baseline on every patch and annotate chosen subsets of pixels with the output of the CNN classifier applying averaging in the overlapping regions. As with our main system, to assess the performance of the baseline, we use a committee of three CNN classifiers at three scales.

\subsubsection* {Nearest neighbor baseline.} 
We have also evaluated a baseline that replaces the learned neural codes with ``hand-crafted'' features. For this, we used SIFT vectors computed over the input $M\times M$ patches as descriptors and use these vectors to perform the nearest-neighbor search in the training dataset. Since SIFT was designed mainly for natural RGB photographs, we evaluate this baseline for the BSDS500 edge detection only.

\subsubsection* {Alternative encoding.} 
Given the impressive results of \cite{Dollar13} on edge detection, we experimented with a variation of our method inspired by their method. We annotate each patch with a long binary vector that looks at the pairs of pixels in the output $N\times N$ patch and assigns it $1$ or $0$ depending whether it belongs to the object segment. We then apply PCA dimensionality reduction to 16 components. More formally, we define the target annotation vector during the CNN training to be:
\begin{equation}
  \B{P} = \PCA ((v_1, v_2, \ldots, v_L)) \, ,
\end{equation}
where $ L = {N^2 \choose 2} $ and $ v_i $ is defined for $ i $-th pair $ (p_l, p_m) $ of pixels in the ground truth segmentation \Seg{\P} and is equal to $ \mathbf 1 \left\{ \Seg{\P}[p_l] = \Seg{\P}[p_m] \right\} $. In the experiments, we observe a small improvement for such alternative encoding.

\subsubsection* {BSDS500 experiments.}
\label{sect:results_bsds500}

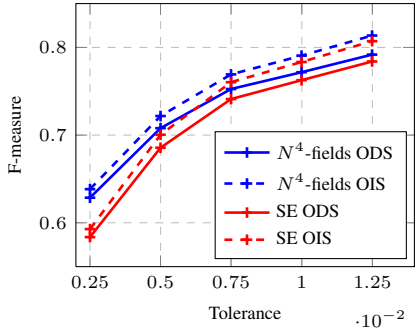
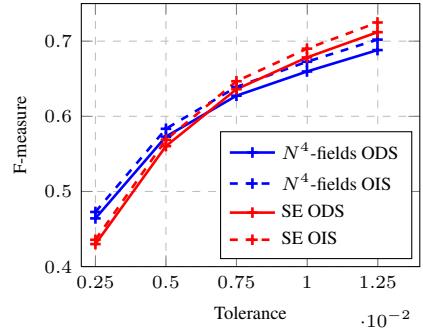
\begin{figure}
  \centering
  \begin{subfigure}[b]{0.5\textwidth}
  	\centering
    \setlength\figureheight{3.5cm}
    \setlength\figurewidth{4.5cm}
%
%
\begin{tikzpicture}[font=\scriptsize]

\begin{axis}[%
width=\figurewidth,
height=\figureheight,
scale only axis,
xmin=0.002,
xmax=0.014,
xlabel={Tolerance},
xmajorgrids,
ymin=0.55,
ymax=0.85,
ylabel={F-measure},
ymajorgrids,
xtick={0.0025,0.005,...,0.015},
legend style={at={($ (1,0) + (-0.1cm,0.1cm) $)},anchor=south east,draw=black,fill=white,legend cell align=left,
grid style={dashed}}
]
\addplot [color=blue,solid,line width=1.0pt,mark=+,mark options={solid}]
  table[row sep=crcr]{
0.0025	0.628617	\\
0.005	0.707917	\\
0.0075	0.752562	\\
0.01	0.771712	\\
0.0125	0.791808	\\
};
\addlegendentry{\NQ-fields ODS};

\addplot [color=blue,dashed,line width=1.0pt,mark=+,mark options={solid}]
  table[row sep=crcr]{
0.0025	0.638373	\\
0.005	0.721768	\\
0.0075	0.769246	\\
0.01	0.790625	\\
0.0125	0.813541	\\
};
\addlegendentry{\NQ-fields OIS};

\addplot [color=red,solid,line width=1.0pt,mark=+,mark options={solid}]
  table[row sep=crcr]{
0.0025	0.583688	\\
0.005	0.685587	\\
0.0075	0.74111	\\
0.01	0.762617	\\
0.0125	0.783965	\\
};
\addlegendentry{SE ODS};

\addplot [color=red,dashed,line width=1.0pt,mark=+,mark options={solid}]
  table[row sep=crcr]{
0.0025	0.592844	\\
0.005	0.700399	\\
0.0075	0.760365	\\
0.01	0.783409	\\
0.0125	0.80715	\\
};
\addlegendentry{SE OIS};

\end{axis}
\end{tikzpicture}%
    \caption{BSDS500 \cite{Arbelaez11}}
    \label{fig:results_bsds500_perf_of_tolerance}
  \end{subfigure}%
  \begin{subfigure}[b]{0.5\textwidth}
  	\centering
    \setlength\figureheight{3.5cm}
    \setlength\figurewidth{4.5cm}
%
%
\begin{tikzpicture}[font=\scriptsize]

\begin{axis}[%
height=\figureheight,
width=\figurewidth,
scale only axis,
xmin=0.002,
xmax=0.014,
xlabel={Tolerance},
xmajorgrids,
ymin=0.4,
ymax=0.75,
ylabel={F-measure},
ymajorgrids,
xtick={0.0025,0.005,...,0.015},
legend style={at={($ (1,0) + (-0.1cm,0.1cm) $)},anchor=south east,draw=black,fill=white,legend cell align=left,
grid style={dashed}}
]
\addplot [color=blue,solid,line width=1.0pt,mark=+,mark options={solid}]
  table[row sep=crcr]{
0.0025	0.464192	\\
0.005	0.573026	\\
0.0075	0.627418	\\
0.01	0.659658	\\
0.0125	0.688118	\\
};
\addlegendentry{\NQ-fields ODS};

\addplot [color=blue,dashed,line width=1.0pt,mark=+,mark options={solid}]
  table[row sep=crcr]{
0.0025	0.472748	\\
0.005	0.583393	\\
0.0075	0.639148	\\
0.01	0.672595	\\
0.0125	0.702049	\\
};
\addlegendentry{\NQ-fields OIS};

\addplot [color=red,solid,line width=1.0pt,mark=+,mark options={solid}]
  table[row sep=crcr]{
0.0025	0.429976	\\
0.005	0.560345	\\
0.0075	0.636067	\\
0.01	0.678334	\\
0.0125	0.711943	\\
};
\addlegendentry{SE ODS};

\addplot [color=red,dashed,line width=1.0pt,mark=+,mark options={solid}]
  table[row sep=crcr]{
0.0025	0.435749	\\
0.005	0.568772	\\
0.0075	0.646657	\\
0.01	0.689998	\\
0.0125	0.724842	\\
};
\addlegendentry{SE OIS};

\end{axis}
\end{tikzpicture}%
    \caption{NYU RGBD \cite{Silberman11}}
    \label{fig:results_nyu_perf_of_tolerance}
  \end{subfigure}
  \caption{Performance scores for different tolerance thresholds (default value is $ \mathbf{0.75 \cdot 10^{-2}} $) used in the BSDS500 benchmark \cite{Arbelaez11}. Algorithms' performance (ODS and OIS measures plotted as {\it dashed} and {\it solid} lines respectively) is going down as the tolerance threshold is decreased. \NQ-fields ({\it blue} lines) handles more stringent thresholds better, which suggests that cleaner edges are produced, as is also evidenced by the qualitative results. See \sect{results} for details.\vspace{-2mm}}
  \label{fig:results_perf_of_tolerance}
\end{figure}

\begin{table}
  \centering
  \begin{subtable}[b]{0.5\textwidth}
  	\centering
    \setlength{\tabcolsep}{4pt}
    \renewcommand{\arraystretch}{1.5}
    \begin{tabular}{l | l | c c c}
      & & ODS & OIS & AP\\
      \hline
      \multirow{10}{*}{\rotatebox[origin=c]{90}{Any}}
      & SIFT + NNB & .59 & .60 & .60\\
      & CNN, central & .72 & .74 & .75\\
      & CNN, patch & .73 & .75 & .74\\
      \cline{2-5}
      & gPb-owt-ucm \cite{Arbelaez11} & .73 & .76 & .73\\
      & SCG \cite{Ren12} & .74 & .76 & .77\\
      & SE-MS, $ T = 4 $ \cite{Dollar13} & .74 & .76 & {\bf .78}\\
      & DeepNet \cite{Kivinen14} & .74 & .76 & .76\\
      & PMI + sPb, MS \cite{Isola14} & .74 & {\bf .77} & {\bf .78}\\
      \cline{2-5}
      & \NQ-fields & {\bf .75} & .76 & .77\\
      & \NQ-fields, AE & {\bf .75} & {\bf .77} & {\bf .78}\\
      \hline
      \multirow{4}{*}{\rotatebox[origin=c]{90}{Consensus}}
      & SE-MS, $ T = 4 $ \cite{Dollar13} & .59 & .62 & .59\\
      & DeepNet \cite{Kivinen14} & .61 & .64 & .61\\
      & PMI + sPb, MS \cite{Isola14} & .61 & {\bf .68} & .56\\
      & \NQ-fields, AE & {\bf .64} & .67 & {\bf .64}\\ 
      \hline
    \end{tabular}
    \caption{BSDS500 \cite{Arbelaez11}}
    \label{tab:results_bsds500}%
    \setlength{\tabcolsep}{1.4pt}
    \renewcommand{\arraystretch}{1.0}
  \end{subtable}%
  \begin{subtable}[b]{0.5\textwidth}
  	\centering
    \setlength{\tabcolsep}{4pt}
    \renewcommand{\arraystretch}{1.5}
    \begin{tabular}{l | c c c}
      & ODS & OIS & AP\\
      \hline
      CNN, central & .60 & .62 & .55\\
      CNN, patch & .58 & .59 & .49\\
      \hline
      gPb \cite{Arbelaez11} & .53 & .54 & .40\\
      SCG \cite{Ren12} & .62 & .63 & .54\\
      SE-MS, $ T = 4 $ \cite{Dollar13} & {\bf .64} & {\bf .65} & {\bf .59}\\
      \hline
      \NQ-fields & .61 & .62 & .56\\
      \NQ-fields, AE & .63 & .64 & .58\\
      \hline
    \end{tabular}
    \caption{NYU RGBD \cite{Silberman11}}
    \label{tab:results_nyu}%
    \setlength{\tabcolsep}{1.4pt}
    \renewcommand{\arraystretch}{1.0}
  \end{subtable}
  \caption{Edge detection results on BSDS500 \cite{Arbelaez11} (both for the original ground-truth annotation and ``consensus'' labels) and NYU RGBD \cite{Silberman11}. Our approach (\NQ-fields) achieves performance which is better or comparable to the state-of-the-art. We also observe that the relative performance of the methods in terms of perceptual quality are not adequately reflected by the standard performance measures.\vspace{-2mm}}
  \label{tab:results}
\end{table}

\begin{figure}
  \centering
  \setlength\figureheight{4.5cm}
  \setlength\figurewidth{8cm}
  \input{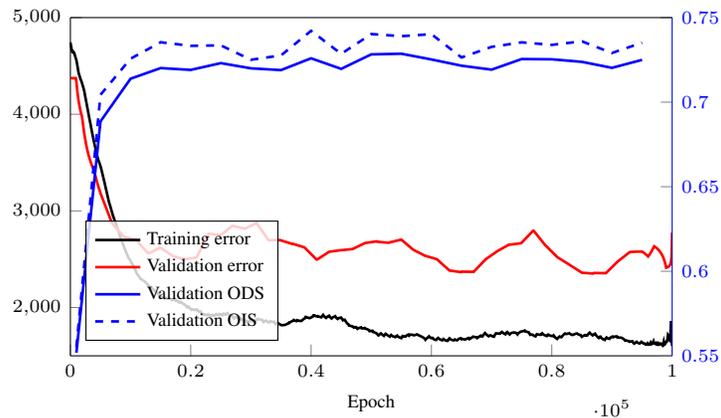}
  \caption{The validation score (average precision) of the full \NQ-fields and error rates (loss) of the underlying CNN measured throughout the training process. The strong correlation between the values suggests the importance of large-scale learning for the good performance of \NQ-fields. This experiment was performed for the BSDS500 edge detection (hold out validation set included 20 images).}
  \label{fig:results_bsds500_evolution}
\end{figure}

The first dataset is Berkley Segmentation Dataset and Benchmark (BSDS500) \cite{Arbelaez11}. It contains 500 color images divided into three subsets: 200 for training, 100 for validation and 200 for testing. Edge detection accuracy is measured using three scores: fixed contour threshold (ODS), per-image threshold (OIS), and average precision (AP) \cite{Arbelaez11,Dollar13}. In order to be evaluated properly, test edges must be thinned to one pixel width before running the benchmark code. We use the non-maximum suppression algorithm from \cite{Dollar13} for that.

\begin{figure}
\centering
\setlength\figurewidth{0.3\textwidth}
\begin{tabular}{ccc}
\includegraphics[width=\figurewidth]{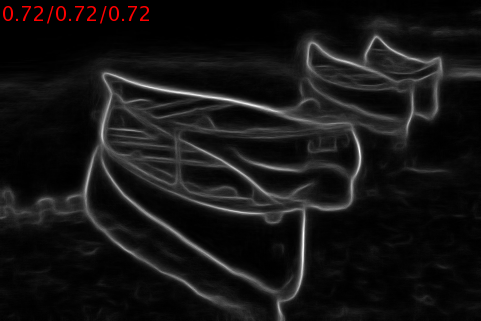}&
\includegraphics[width=\figurewidth]{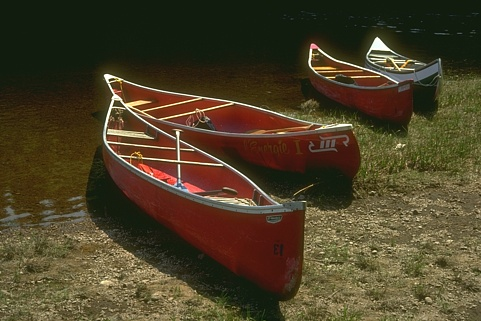}&
\includegraphics[width=\figurewidth]{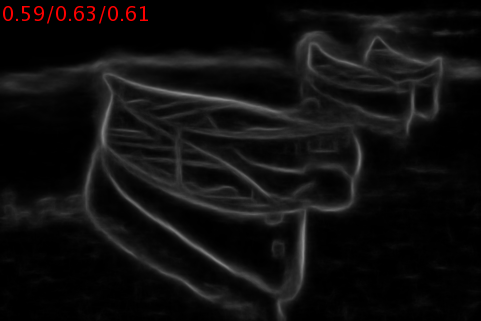}\\
\includegraphics[width=\figurewidth]{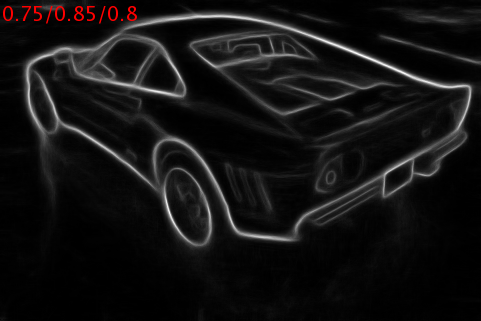}&
\includegraphics[width=\figurewidth]{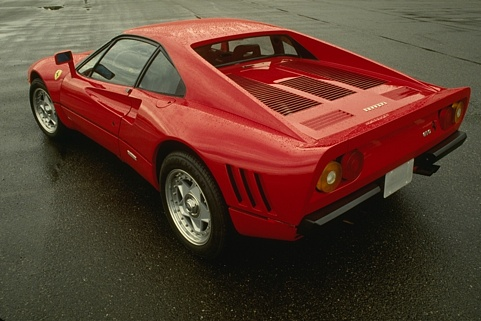}&
\includegraphics[width=\figurewidth]{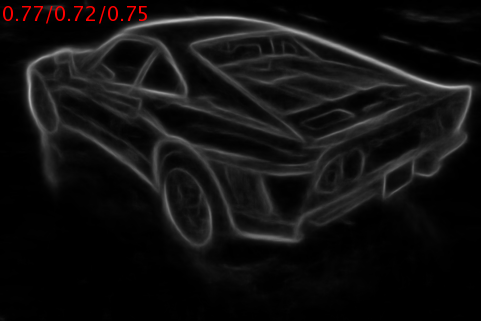}\\
\includegraphics[width=\figurewidth]{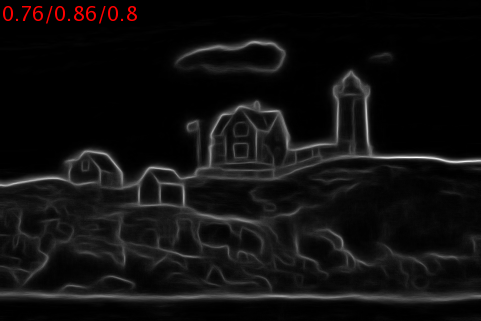}&
\includegraphics[width=\figurewidth]{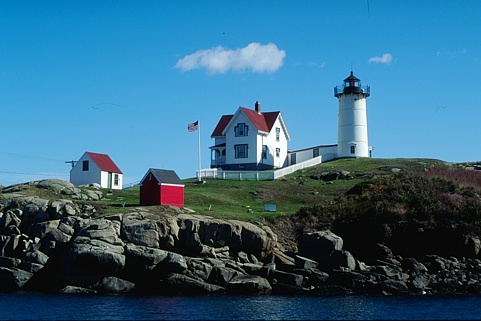}&
\includegraphics[width=\figurewidth]{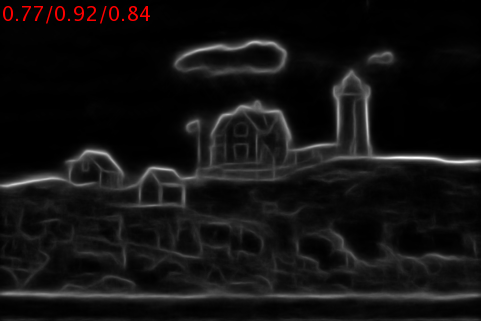}\\
\NQ-fields & Input & Structured Edge~\cite{Dollar13}\vspace{-2mm}
\end{tabular}
\caption{Representative results on the BSDS500 dataset. For comparison, we give the results of the best previously published method \cite{Dollar13}. The red numbers correspond to Recall/Precision/F-measure. We give two examples where \NQ-fields perform better than \cite{Dollar13}, and one example (bottom row) where \cite{Dollar13} performs markedly better according to the quantitative measure.\vspace{-2mm}}
\label{fig:bsds_results}
\end{figure}

In general, \NQ-fields perform similarly to the best previously published methods~\cite{Dollar13,Kivinen14,Isola14}. In particular, the full version of the system (the committee of three \NQ-fields applied at three scales) matches the performance of the mentioned algorithms, with the alternative encoding performing marginally better. Following \cite{Hou13} in order to account for the inherent problems of the dataset we also test our approach against the so-called ``consensus'' subset of the ground-truth labels. Within this setting our method significantly outperforms other algorithms in terms of ODS and AP (\tab{results}~-~left).

The benchmark evaluation procedure does not perform strict comparison of binary edge masks but rather tries to find the matching between pixels within certain tolerance level and then analyzes unmatched pixels \cite{Arbelaez11}. We observed that the default distance matching tolerance threshold, while accounting for natural uncertainty in the exact position of the boundary, often ignores noticeable and unnatural segmentation mistakes such as spurious boundary pixels. Therefore, in addition to the accuracy evaluated for the default matching threshold, we report results for more stringent thresholds (\fig{results_bsds500_perf_of_tolerance}-left).

It is also useful to investigate how successful is the deep learning, and what is its role within the \NQ-fields. It is insightful to see whether the outputs of the CNN within the \NQ-fields, i.e.\  $\CNN(\P)$ are reasonably close to the codes $\PCA(\A{\P})$ that were used as target during the learning. To show this, in \fig{nearest} we give several representative results of the nearest neighbor searches where different types of codes are used on the query and on the dictionary dataset sides (alongside the corresponding patches). It can be seen, that there are very accurate matches (in terms of similarity between true annotations) between PCA codes on both sides, and reasonably good matches between CNN codes on both sides. However, when matching the CNN code of an input patch to PCA codes on the dataset side the results are poor. This is especially noticeable for patches without natural boundaries in them as we force our neural network to map all such patches into one point (empty annotation is always encoded with the same vector). This qualitative performance results in a notoriously bad quantitative performance of the system that uses such matching (from the CNN codes in the test image to the PCA codes in the training dataset).

While CNN is clearly unable to learn to reproduce the target codes closely, there is still a strong correlation between the training error (the value of the loss function within the CNN) and the performance of the \NQ-fields (\fig{results_bsds500_evolution}). The efficiency of the learned codes and its importance for the good performance of \NQ-fields is also highlighted by the fact that the nearest neighbor baseline using SIFT codes performs very poorly. Thus, optimizing the loss functions introduced above really makes edge maps produced by our algorithm agree with ground truth annotations.

\subsubsection* {NYU RGBD experiments.}
\label{sect:results_nyu}

\begin{figure}
\setlength\figurewidth{0.17\textwidth}
\newcommand{\resrow}[1]{%
\includegraphics[width=\figurewidth]{Images/nyu_results/#1_in.png}&
\includegraphics[width=\figurewidth]{Images/nyu_results/#1_out.png}&
\includegraphics[width=\figurewidth]{Images/nyu_results/#1_cnn_patch.png}&
\includegraphics[width=\figurewidth]{Images/nyu_results/#1_cnn_central.png}&
\includegraphics[width=\figurewidth]{Images/nyu_results/#1_dollar.png}%
}
\centering
\vspace{-6mm}
\begin{tabular}{ccccc}
\resrow{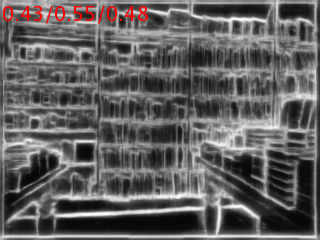}\\
\resrow{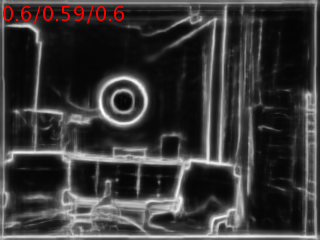}\\
\resrow{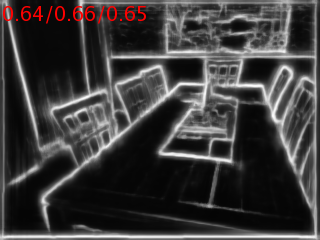}\\
Input & \NQ-fields & CNN, patch & CNN, central & SE~\cite{Dollar13}
\end{tabular}
\caption{Results on the NYU RGBD dataset. For comparison, we give the results of the best previously published method \cite{Dollar13} and the CNN baseline. We show a representative result where the \NQ-fields perform better (top), similarly (middle), or worse (bottom) than the baseline, according to the numberic measures shown in red (recall/precision/F-measure format).
We argue that the numerical measures do not adequately reflect the relative perceptual performance of the methods (see also \cite{SupMat} for more examples). }
\label{fig:nyu_results}
\end{figure}

We now present results for the NYU Depth dataset (v2) \cite{Silberman11}. It contains 1,449 RGBD images with corresponding semantic segmentations. Ren and Bo \cite{Ren12} developed an utility script which translates the data into BSDS500 format thus eliminating any need for the adaptation of our pipeline for the new dataset. To make the comparison with the previous approaches easier we use the training/testing split proposed by \cite{Ren12}. The CNN architecture stays the same except for the number of input channels which is now equal to 4 instead of 3.

Once again we use the BSDS500 benchmark code to assess the performance of different algorithms. The results are summarized in \tab{results}-right. Our approach almost ties the state-of-the-art method by \cite{Dollar13}. However, just like in the case of the BSDS500 dataset this difference in scores may be due to the peculiarity of the benchmark described in the previous section. Indeed, \fig{results_nyu_perf_of_tolerance}-right shows that for smaller values of matching thresholds, \NQ-fields match or even outperform the accuracy of Structured Edge detector \cite{Dollar13}. 

\subsubsection* {Note on the quantitative performance.} During the experiments, we observed a clear disconnect between the relative performance of the methods according to the quantitative measures, and according to the actual perceptual quality, in particular on the NYU RGBD dataset (\fig{nyu_results}). We provide extended uniformly-sampled qualitative results in the supplementary materials~\cite{SupMat}.     

\subsubsection* {DRIVE dataset.}
\label{sect:results_drive}

\begin{figure}
\centering
\vspace{-6mm}
\setlength\figurewidth{0.22\textwidth}
\begin{tabular}{cccc}
\includegraphics[width=\figurewidth]{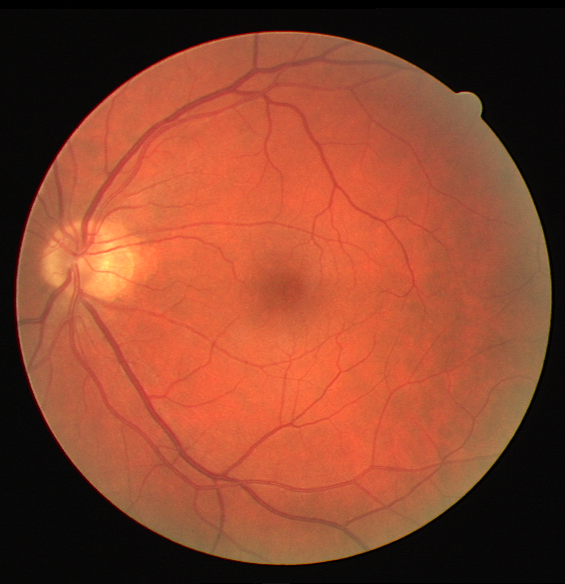}&
\includegraphics[width=\figurewidth]{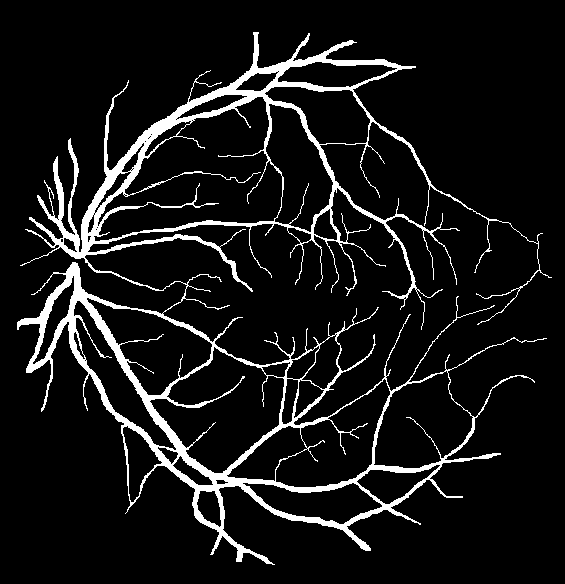}&
\includegraphics[width=\figurewidth]{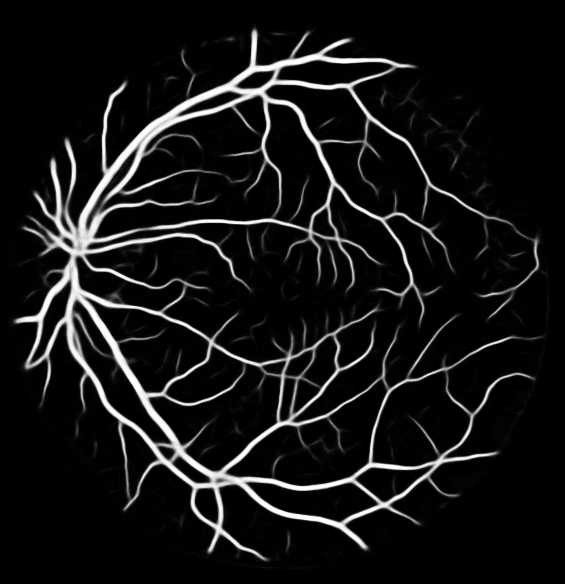}&
\includegraphics[width=\figurewidth]{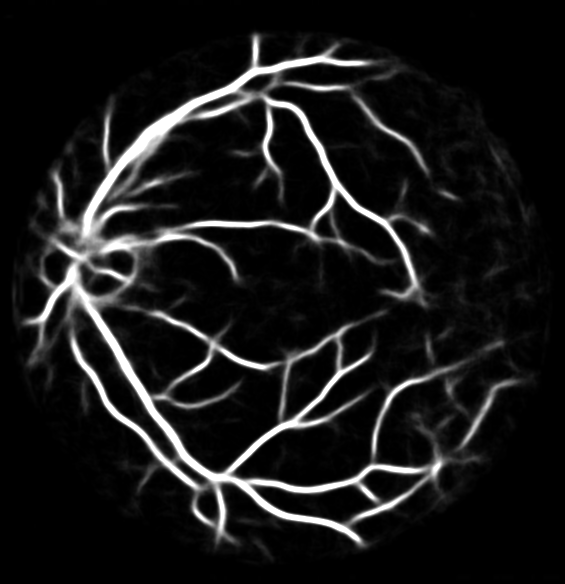}\\
\includegraphics[width=\figurewidth]{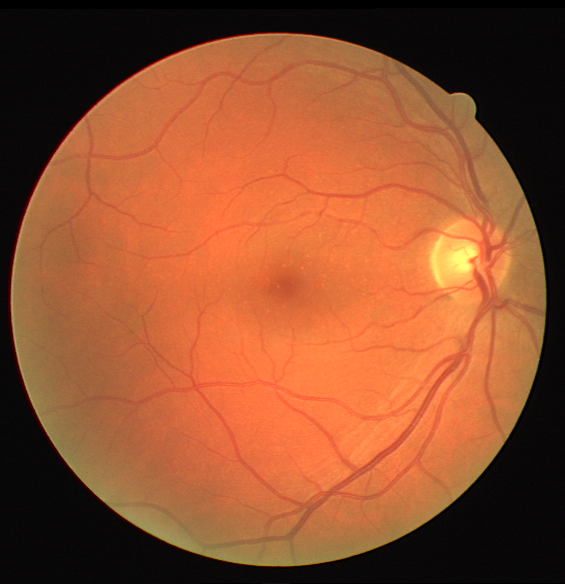}&
\includegraphics[width=\figurewidth]{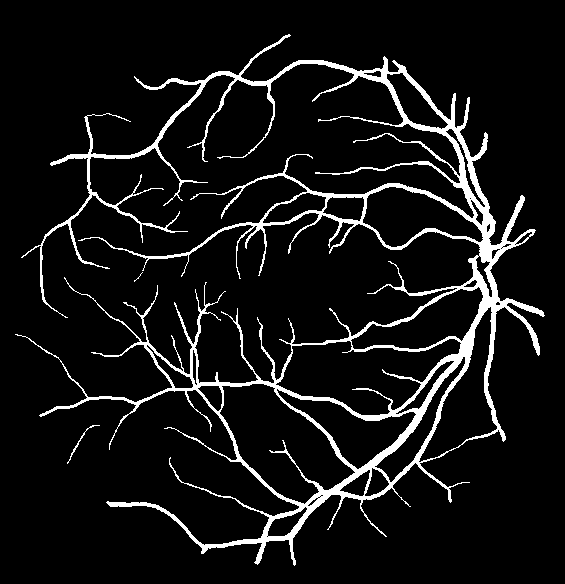}&
\includegraphics[width=\figurewidth]{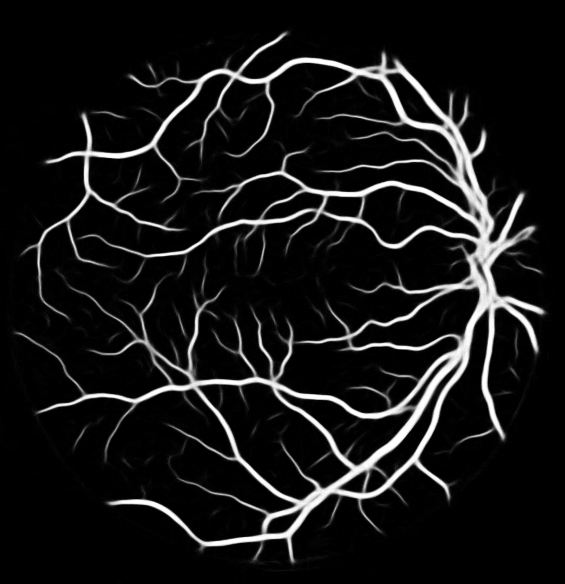}&
\includegraphics[width=\figurewidth]{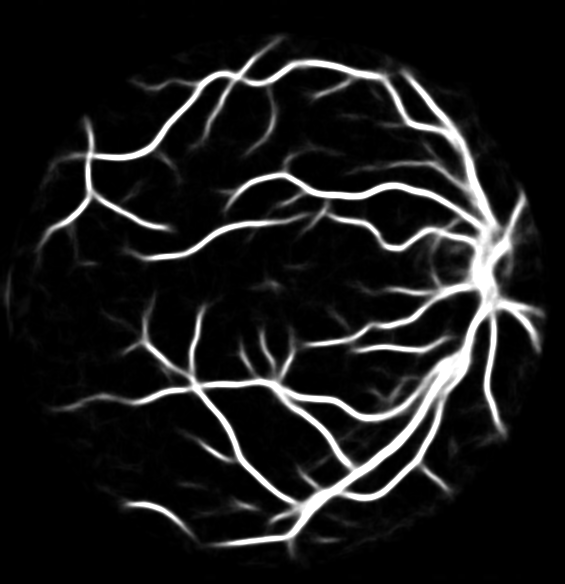}\\
Input & Expert annotation & \NQ-fields & SE~\cite{Dollar13} \vspace{-2mm}
\end{tabular}
\caption{Representative results on the DRIVE dataset. A close match to the human expert annotation is observed.\vspace{-2mm}}
\label{fig:drive_results}
\end{figure}

\begin{figure}
  \centering
  \setlength\figureheight{4.5cm}
  \setlength\figurewidth{8cm}
  \input{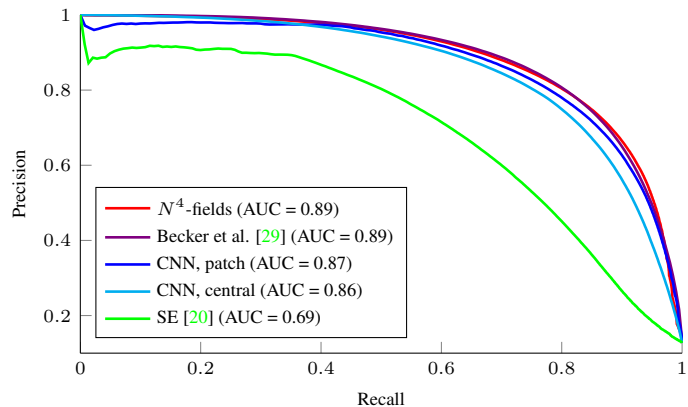}\vspace{-2mm}
  \caption{Results for the DRIVE dataset \cite{Staal04} in the form of the recall/precision curves. Our approach matches the performance of the current state-of-the-art method by Becker et al \cite{Becker13} and performs much better than baselines and \cite{Dollar13}. \vspace{-2mm}}
  \label{fig:results_drive_pr}
\end{figure} 

In order to demonstrate wide applicability of our method we test it on the DRIVE dataset \cite{Staal04} of the micrographs obtained within the diabetic retinopathy screening program. There are 40 images of size 768~$ \times $~584 divided evenly into a training and a test set. Ground truth annotations include manually segmented vasculature as well as ROI masks.

We use exactly the same CNN architecture as in the BSDS500 experiment. Without any further tuning our system achieves state-of-the-art performance comparable to the algorithm proposed by Becker et al. \cite{Becker13}. Precision/recall curves for both approaches as well as for the baseline neural networks and \cite{Dollar13} are shown in \fig{results_drive_pr}. Notably, there is once again a clear advantage over the CNN classifier baselines. Poor performance of \cite{Dollar13} is likely to be due to the use of default features that are not suitable for this particular imaging modality. This provides extra evidence for the benefits of fully data-driven approach.

\section{Conclusion}
\label{sect:discussion}

We have presented a new approach to machine-learning based image processing. We have demonstrated how convolutional neural networks can be efficiently combined with the nearest neighbor search, and how such combination can improve over the performance of standalone CNNs in the situation when CNN training underfits due to the problem complexity. State-of-the-art results are demonstrated for natural edge detection in RGB and RGBD images, as well as for thin object (vessel) segmentation. To the best of our knowledge, these are the first state-of-the-art results for natural edge detection obtained with deep learning. Compared to the structured forests method \cite{Dollar13}, the proposed approach is slower, but can be adapted to new domains (e.g.\ micrographs) without any retuning.

The future work may concern the fact that we use a PCA compression to define the target output during the CNN training. A natural idea is then to learn some non-linear transformation in the label space in parallel to the CNN training on the image patch input, so that to minimize the gap between the neural codes of the input patches and the target annotations, which remains considerable in our experiments. It remains to be seen whether minimizing this gap further will bring the improvement to the overall performance of the system.

\clearpage

\bibliographystyle{splncs}
\bibliography{references}

\begin{thebibliography}{10}

\bibitem{LeCun89}
LeCun, Y., Boser, B.E., Denker, J.S., Henderson, D., Howard, R.E., Hubbard,
  W.E., Jackel, L.D.:
\newblock Handwritten digit recognition with a back-propagation network.
\newblock In: NIPS. (1989)  396--404

\bibitem{Krizhevsky12}
Krizhevsky, A., Sutskever, I., Hinton, G.:
\newblock {Imagenet classification with deep convolutional neural networks}.
\newblock Advances in Neural Information \ldots (2012)  1--9

\bibitem{Ciresan12}
Ciresan, D.C., Giusti, A., Gambardella, L.M., Schmidhuber, J.:
\newblock Deep neural networks segment neuronal membranes in electron
  microscopy images.
\newblock In: NIPS. (2012)  2852--2860

\bibitem{Arbelaez11}
Arbel\'{a}ez, P., Maire, M., Fowlkes, C., Malik, J.:
\newblock {Contour detection and hierarchical image segmentation.}
\newblock IEEE transactions on pattern analysis and machine intelligence
  \textbf{33}(5) (May 2011)  898--916

\bibitem{Silberman11}
Silberman, N., Fergus, R.:
\newblock Indoor scene segmentation using a structured light sensor.
\newblock In: ICCV Workshops, IEEE (2011)  601--608

\bibitem{Staal04}
Staal, J., Abràmoff, M.D., Niemeijer, M., Viergever, M.A., van Ginneken, B.:
\newblock Ridge-based vessel segmentation in color images of the retina.
\newblock IEEE Trans. Med. Imaging \textbf{23}(4) (2004)  501--509

\bibitem{Egmont02}
Egmont-Petersen, M., de~Ridder, D., Handels, H.:
\newblock Image processing with neural networks -— a review.
\newblock Pattern recognition \textbf{35}(10) (2002)  2279--2301

\bibitem{Mnih10}
Mnih, V., Hinton, G.E.:
\newblock Learning to detect roads in high-resolution aerial images.
\newblock In: Computer Vision--ECCV 2010.
\newblock Springer (2010)  210--223

\bibitem{Schulz12}
Schulz, H., Behnke, S.:
\newblock Learning object-class segmentation with convolutional neural
  networks.
\newblock In: 11th European Symposium on Artificial Neural Networks (ESANN).
  Volume~3. (2012)

\bibitem{Kivinen14}
Kivinen, J.J., Williams, C.K.I., Heess, N.:
\newblock Visual boundary prediction: A deep neural prediction network and
  quality dissection.
\newblock In: AISTATS. (2014)  512--521

\bibitem{Ranzato10}
Ranzato, M., Hinton, G.E.:
\newblock Modeling pixel means and covariances using factorized third-order
  boltzmann machines.
\newblock In: CVPR. (2010)  2551--2558

\bibitem{Farabet13}
Farabet, C., Couprie, C., Najman, L., LeCun, Y.:
\newblock Learning hierarchical features for scene labeling.
\newblock Pattern Analysis and Machine Intelligence, IEEE Transactions on
  \textbf{35}(8) (2013)  1915--1929

\bibitem{Jain07}
Jain, V., Murray, J.F., Roth, F., Turaga, S.C., Zhigulin, V.P., Briggman, K.L.,
  Helmstaedter, M., Denk, W., Seung, H.S.:
\newblock Supervised learning of image restoration with convolutional networks.
\newblock In: ICCV. (2007)  1--8

\bibitem{Jain08}
Jain, V., Seung, H.S.:
\newblock Natural image denoising with convolutional networks.
\newblock In: NIPS. (2008)  769--776

\bibitem{Burger12}
Burger, H.C., Schuler, C.J., Harmeling, S.:
\newblock Image denoising: Can plain neural networks compete with bm3d?
\newblock In: Computer Vision and Pattern Recognition (CVPR), 2012 IEEE
  Conference on, IEEE (2012)  2392--2399

\bibitem{Chopra05}
Chopra, S., Hadsell, R., LeCun, Y.:
\newblock Learning a similarity metric discriminatively, with application to
  face verification.
\newblock In: Computer Vision and Pattern Recognition, 2005. CVPR 2005. IEEE
  Computer Society Conference on. Volume~1., IEEE (2005)  539--546

\bibitem{Dabov08}
Dabov, K., Foi, A., Katkovnik, V., Egiazarian, K.:
\newblock Image restoration by sparse 3d transform-domain collaborative
  filtering.
\newblock In: Electronic Imaging 2008, International Society for Optics and
  Photonics (2008)  681207--681207

\bibitem{Criminisi04}
Criminisi, A., P{\'e}rez, P., Toyama, K.:
\newblock Region filling and object removal by exemplar-based image inpainting.
\newblock Image Processing, IEEE Transactions on \textbf{13}(9) (2004)
  1200--1212

\bibitem{Freeman00}
Freeman, W.T., Pasztor, E.C., Carmichael, O.T.:
\newblock Learning low-level vision.
\newblock International Journal of Computer Vision \textbf{40}(1) (2000)
  25--47

\bibitem{Dollar13}
Doll\'ar, P., Zitnick, C.L.:
\newblock Structured forests for fast edge detection.
\newblock In: ICCV. (2013)

\bibitem{Hinton12}
Hinton, G.E., Srivastava, N., Krizhevsky, A., Sutskever, I., Salakhutdinov, R.:
\newblock Improving neural networks by preventing co-adaptation of feature
  detectors.
\newblock CoRR \textbf{abs/1207.0580} (2012)

\bibitem{Zeiler12}
Zeiler, M.D., Fergus, R.:
\newblock {Visualizing and Understanding Convolutional Networks}.
\newblock (2012)

\bibitem{Sermanet14}
Sermanet, P., Eigen, D.:
\newblock {OverFeat : Integrated Recognition , Localization and Detection using
  Convolutional Networks arXiv : 1312 . 6229v3 [ cs . CV ] 14 Jan 2014}.
\newblock  1--16

\bibitem{Vedaldi08vlfeat}
Vedaldi, A., Fulkerson, B.:
\newblock {VLFeat}: An open and portable library of computer vision algorithms.
\newblock \url{http://www.vlfeat.org/} (2008)

\bibitem{Ren12}
Xiaofeng, R., Bo, L.:
\newblock Discriminatively trained sparse code gradients for contour detection.
\newblock In Bartlett, P., Pereira, F., Burges, C., Bottou, L., Weinberger, K.,
  eds.: Advances in Neural Information Processing Systems 25.
\newblock (2012)  593--601

\bibitem{Isola14}
Isola, P., Zoran, D., Krishnan, D., Adelson, E.H.:
\newblock Crisp boundary detection using pointwise mutual information.
\newblock In: ECCV. (2014)

\bibitem{Hou13}
Hou, X., Yuille, A., Koch, C.:
\newblock Boundary detection benchmarking: Beyond f-measures.
\newblock In: Computer Vision and Pattern Recognition, 2013. CVPR'13. Volume
  2013., IEEE (2013)  1--8

\bibitem{SupMat}
Ganin, Y., Lempitsky, V.:
\newblock Online supplementary material for the article ``{$N^4$-Fields}:
  {N}eural {N}etwork {N}earest {N}eighbor fields for image transforms''.
\newblock \url{http://sites.skoltech.ru/compvision/projects/n4/}

\bibitem{Becker13}
Becker, C.J., Rigamonti, R., Lepetit, V., Fua, P.:
\newblock Supervised feature learning for curvilinear structure segmentation.
\newblock In Mori, K., Sakuma, I., Sato, Y., Barillot, C., Navab, N., eds.:
  MICCAI (1). Volume 8149 of Lecture Notes in Computer Science., Springer
  (2013)  526--533

\end{thebibliography}
\end{document}